%% file: ms.tex
\documentclass[sigconf]{acmart}

\usepackage[english]{babel}
\usepackage{blindtext}

\renewcommand\footnotetextcopyrightpermission[1]{} 
\setcopyright{none}

\acmDOI{}

\acmISBN{}

\acmPrice{}

\usepackage{microtype}
\usepackage{graphicx}
\usepackage{subfigure}
\usepackage{graphicx}
\usepackage{booktabs} 
\usepackage{listings}
\usepackage{tabularx}
\graphicspath{ {fig/} }
\usepackage{hyperref}

\usepackage{amsmath}
\usepackage{amsfonts}
\usepackage{amssymb}
\usepackage{mathtools}
\usepackage[linesnumbered,vlined,ruled,commentsnumbered]{algorithm2e}
\SetKwInput{KwIn}{Input}%
\SetKwInput{KwOut}{Output}%
\SetStartEndCondition{ }{}{}%
\SetKwProg{Fn}{def}{\string:}{}
\SetKwFunction{Range}{range}
\SetKw{KwTo}{in}\SetKwFor{For}{for}{\string:}{}%
\SetKwIF{If}{ElseIf}{Else}{if}{:}{elif}{else:}{}%
\SetKwFor{While}{while}{:}{fintq}%
\SetKw{Break}{break}%
\SetKw{Return}{return}%
\SetKw{Append}{append}%
\SetKw{Sort}{sort}%
\SetKwFunction{Range}{range}%
\usepackage[noend]{algpseudocode}

\usepackage{amsfonts}

\begin{document}
\title{Anomaly Detection for Water Treatment System based on  Neural Network with Automatic Architecture Optimization}




\author{Dmitry Shalyga}
\affiliation{%
\institution{Kaspersky Lab}
}
 \email{Dmitry.Shalyga@kaspersky.com}

\author{Pavel Filonov}
\affiliation{%
\institution{Kaspersky Lab}
}
 \email{Pavel.Filonov@kaspersky.com}

\author{Andrey Lavrentyev}
\affiliation{%
\institution{Kaspersky Lab}
}
 \email{Andrey.Lavrentyev@kaspersky.com}

\renewcommand{\shortauthors}{Shalyga.et al.}
\renewcommand{\shorttitle}{ Anomaly detection for  SWaT }

\begin{abstract}
We continue to develop our neural network (NN) based forecasting approach to anomaly detection (AD) using the Secure Water Treatment (SWaT) industrial control system (ICS) testbed dataset. We propose genetic algorithms (GA) to find the best NN architecture for a given dataset, using the NAB metric to assess the quality of different architectures. The drawbacks of the F1-metric are analyzed. Several techniques are proposed to improve the quality of AD: exponentially weighted smoothing, mean p-powered error measure, individual error weight for each variable, disjoint prediction windows. Based on the techniques used, an approach to anomaly interpretation is introduced.
\end{abstract}

\maketitle

\input{intro}
\input{solution}
\input{arch_search}
\input{detection_metrics}
\input{conclusions}

 \input{acknowledgement}

\appendix
\include{appendix_a}

\bibliographystyle{ACM-Reference-Format}
\bibliography{ms}

\end{document}

%% file: intro.tex
\section{Introduction}
\label{intro}

An ICS generates a large amount of data. In this paper, we refer to each data variable (i.e., sensor or actuator value, setpoint, etc.) as a "tag". Taken together, the data can be represented as a multivariate time series. Any anomalies present in such data represent deviations in industrial processes, which may have different causes, such as equipment faults, the human factor, cyberattacks ~\cite{stuxnet}, ~\cite{ukr}, ~\cite{lessons}. Early anomaly detection and anomaly interpretation can help to prevent serious consequences. 

There are numerous methods of AD, such as LSTM-based forecasting ~\citep{KL16} and encoder-decoder ~\cite{Malhotra16}, PCA, DPCA, FDA, DFDA, CVA, PLS ~\citep{FDDinIS}, clustering based ~\cite{TEP_DoS_15}, change point detection ~\cite{ChPoint13}, one-class SVM and segmentation ~\cite{Petrol15}, process invariants ~\cite{ProcInv16}.

Published results are invariably based on some dataset. In the case of ICS, there are relatively few publicly available datasets. We have previously published papers on two datasets, which we used to test our NN-based forecasting approach to AD ~\citep{KL16}, ~\citep{icml2017}. In this work, we use the Secure Water Treatment ~\cite{SWaT_download} dataset generated by a testbed ICS with real-world equipment ~\cite{SWaT}. The dataset includes 51 tags (25 sensors and 26 actuators) collected during two sessions: 7 days under normal operating conditions and 6 days during which 34 attacks that had physical impacts were carried out.  

Other researchers have proposed several approaches to AD in the SWaT dataset: detecting anomalies in a single phase of the industrial process ~\cite{GohRNN}, detecting single-stage multipoint attacks \cite{single_stage}, an invariant-based approach in ~\cite{dataset_ref1} and ~\cite{inv_based}. The first two methods perform well on the attacks that have been studied, but they cannot detect multi-stage attacks and don't provide an anomaly diagnosis, i.e., they do not provide the means of identifying the variable attacked. The last method analyzes only a few tags and can consequently only detect attacks that modify those tags.

In our previous work we manually selected NN architecture based on LSTM or GRU RNN. In this paper, we introduce an approach based on using GA to find the best NN architecture for a given dataset. The best NN architecture found for the SWaT dataset includes encoder, analyzer and decoder blocks, with their hyperparameters optimized for the dataset. 

To further improve the quality of detection, we introduce several useful techniques: exponentially weighted moving average smoothing (used in our previous works), mean p-powered error measure, individual error weight for each tag values, disjoint prediction windows. These techniques allow us to reduce the false-negative rate, balance false positives, and tune the detector to dataset characteristics such as tag value noise and predictability. 

To achieve early detection, we use the NAB metric  ~\cite{NAB}, which is sensitive to detection delays and anomaly duration. 

The proposed approach allows for anomaly interpretation: we can automatically define a list of tags that were potentially attacked for every anomaly detected. This is very helpful in situations that involve multi-stage multipoint attacks.


%% file: solution.tex
\section{Anomaly Detection}
\label{solution}

As a train dataset we use  the part of the SWaT dataset generated under normal operating conditions.
The remaining part of dataset, generated under 34 attack scenarios, is used as a test dataset.
Our approach involves re-interpolating tag values to a uniform time grid. However, in the SWaT dataset, values are already given on a uniform time grid with an interval of one second.
Before applying machine learning methods, we scaled the original data so that every tag value had a mean value of 0 and a standard deviation of 1.

\subsection{Detection Method}

\begin{figure}[h]
\includegraphics[width=\columnwidth]{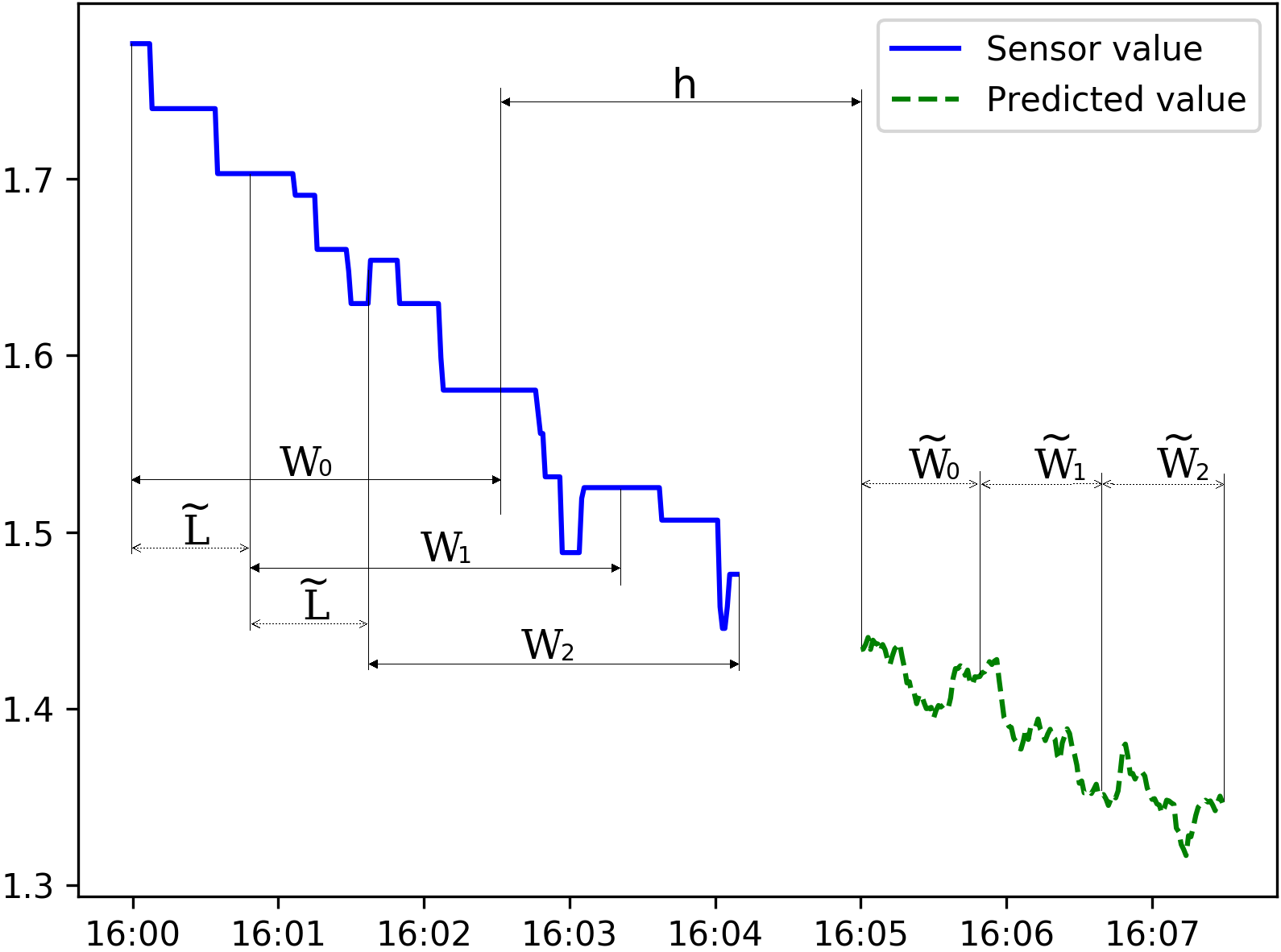}
\caption{Disjoint prediction window for the AIT501 tag with input window $L = 150$, forecast horizon $h = 150$ and forecast length $\tilde{L} = 50$.}
\label{fig:shifted_forecast}
\end{figure}

Denote the number of values (dimensions) for a single time point as $m$.
For the SWaT dataset, it is equal to the number of sensors and actuators, i.e., $m=51$.
For each time point $t \in \mathbb{N}$, let
\begin{equation}
\mathbf{x_t} = (x_{t1}, ..., x_{ti}, ..., x_{tm}), i = \overline{1,m}, x_{ti}\in \mathbb{R}
\end{equation}
where $x_{ti}$ is the value of tag $i$ at timepoint $t$.
Denote the length of the input window as $L \in \mathbb{N}$ and the length of the forecast window as $\tilde{L} \in \mathbb{N}$.
Then, given a total of $S$ timepoints in the dataset, we have $K = \left \lfloor{(S - L - h) / \tilde{L}}\right \rfloor$ input and predicted time windows, where $h \in \mathbb{N}$ is the forecasting horizon (see Figure ~\ref{fig:shifted_forecast}).
In this paper, we assume $h \geq 0$.
Denote the input time window starting with timepoint $t$ and having the length of $L$ points as
\begin{equation}W_k=(\mathbf{x_t}, \mathbf{x_{t+1}}, ..., \mathbf{x_{t+L}}), t=k\tilde{L}, k = \overline{0, K-1}\end{equation}
and the corresponding predicted window as
\begin{equation}\label{eq:pred_window}
\tilde{W}_k=(\mathbf{\tilde{x}_{\tilde{t}}}, \mathbf{\tilde{x}_{\tilde{t}+1}}, ..., \mathbf{\tilde{x}_{\tilde{t}+\tilde{L}}}), \tilde{t}=L + h + k\tilde{L}
\end{equation}

\begin{figure}[h]
\includegraphics[width=\columnwidth]{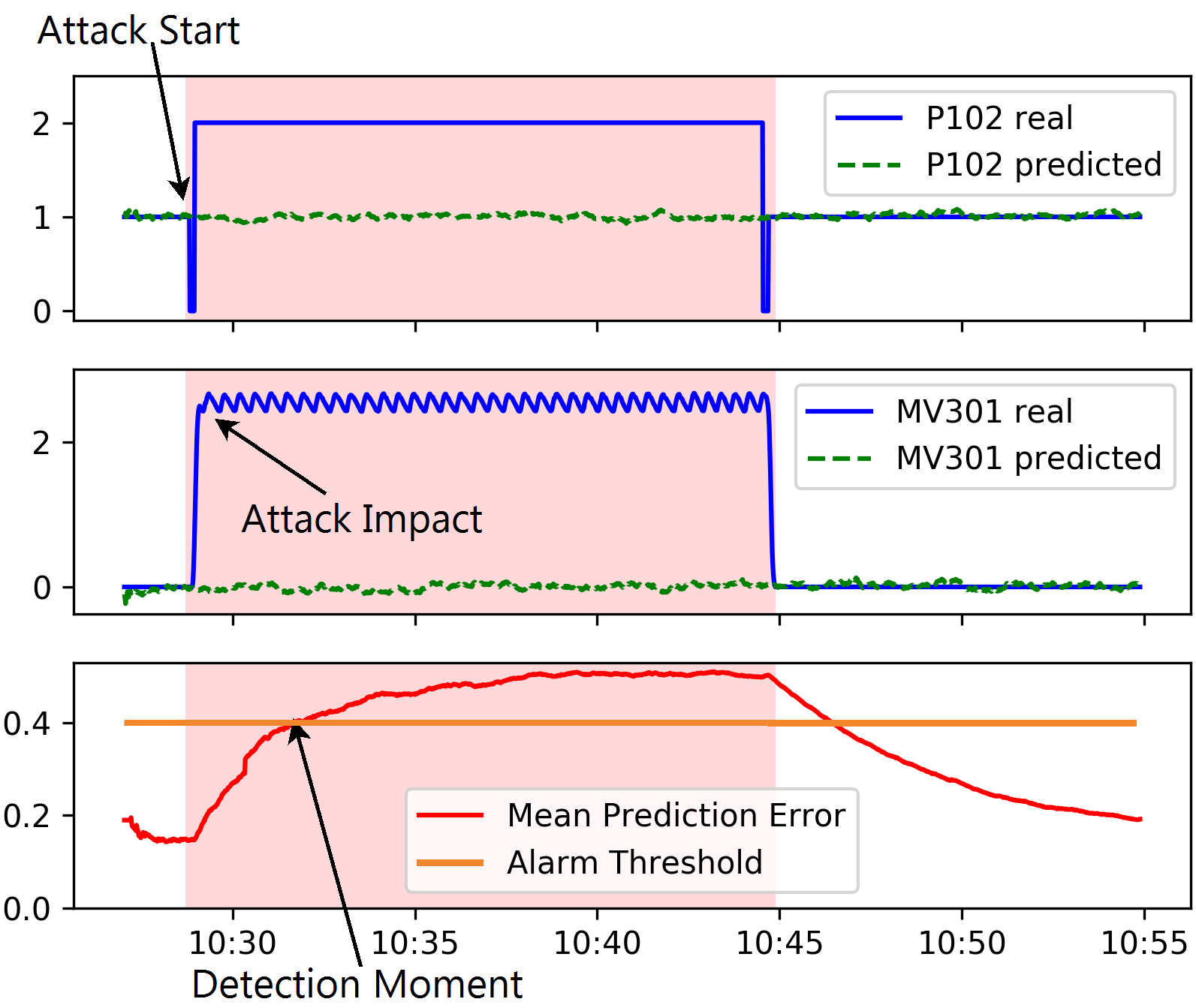}
\caption{Attack detection on the SWaT dataset.}
\label{fig:attack}
\end{figure}

Using a neural network, we approximate a forecasting function $f$ such that
\begin{equation}\tilde{W}_k = f(W_k), f: \mathbb{R}^{L\times m} \to \mathbb{R}^{\tilde{L}\times m}\end{equation}
Given the system's state for time window $W_k$, the model forecasts every tag's value for every second of time window $\tilde{W}_k$.
This means that, according to \eqref{eq:pred_window}, we have calculated forecast values $\mathbf{\tilde{x}_t}$ for each timepoint except the first $L + h$.
For consistency, assume
\begin{equation}\mathbf{\tilde{x}_t} = \mathbf{x_t}, t = \overline{0, L+h-1}\end{equation}
For each timepoint $t$, we calculate the mean error
\begin{equation}M_t^{(0)} = \frac{1}{m}\sum_{i=1}^{m}|\tilde{x}_{ti} - x_{ti}|\end{equation}
Let the threshold value $T$ be the 99th percentile of mean error $M$ on a training set.
Then, for each timepoint $t$, we report an anomaly if $M_t^{(0)} \geq T$ (Figure ~\ref{fig:attack}).

We also use some helpful techniques to improve the quality of anomaly detection.

\subsubsection{Exponentially Weighted Smoothing}
\begin{figure}[h]
\includegraphics[width=\columnwidth]{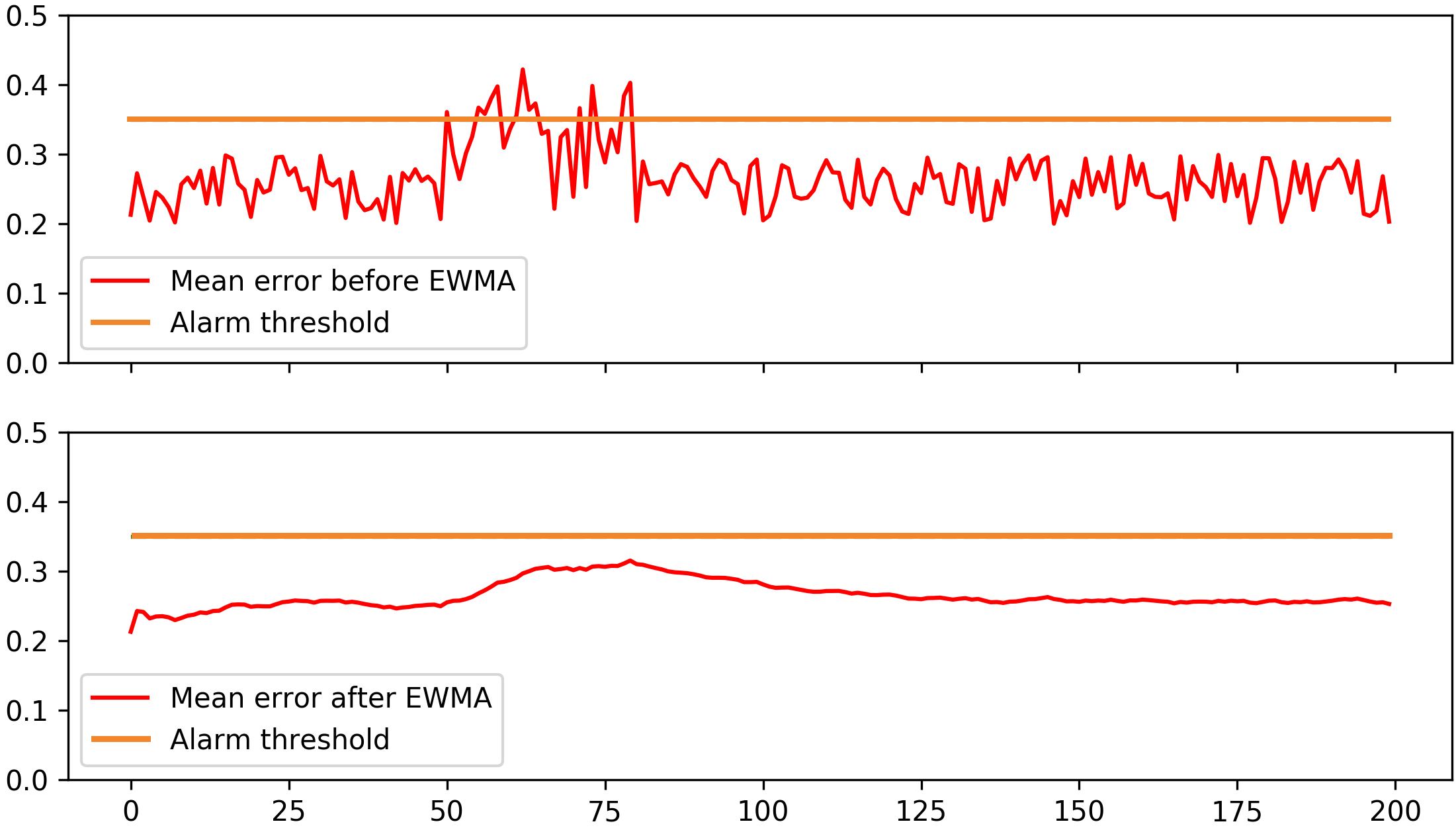}
\caption{Before exponential smoothing, the system report is false positive, and after smoothing with $\alpha = 0.067$ that problem is gone}
\label{fig:ewma}
\end{figure}

The first technique is to use exponentially weighted moving average smoothing:
\begin{equation}M_t^{(1)} = \alpha M_t^{(0)} + (1 - \alpha)M_{t-1}^{(1)}, \end{equation}
\[M_0^{(1)}=0, \alpha=1-\exp^{\frac{\ln(0.5)}{H}}\]
where $H$ is the half-life period.
This method allows us to achieve a considerably lower false-positive rate (Figure ~\ref{fig:ewma}) at the cost of a slightly later anomaly detection.
During our experiments, we discovered that the half-life period $H$ should be equal to the forecast length $\tilde{L}$.

\subsubsection{Mean p-Powered Error}
The second technique is to use power for mean error calculation:
\begin{equation}M_t^{(2)} = \frac{1}{m}\sum_{i=1}^{m}|\tilde{x}_{ti} - x_{ti}|^p\end{equation}
Here $p$ is a parameter of the detection system.
Note that with $p = 2$, this will calculate the \textit{mean squared error}.
The idea of raising the error value to a power greater than 2 is to enable detection of short-term anomalies.
If one of them has an exceptional value, it might be lost after mean error calculation and exponential smoothing, resulting in higher false negative rate.
At the same time, if $p$ is too high, we will only notice outliers, increasing the false positive rate.
This means that we need to choose such $p$ that is not too small (to reduce the false negative rate) but not too high (to avoid increasing the false positive rate).
Based on our experiments, we found the best value to be $p = 6$.

\subsubsection{Weighted p-Powered Error}
There is noise in every real-world system.
And some tags are easier to forecast (strongly periodic or even constant processes).
If such a tag has an unusual value, this can, with a higher degree of confidence, be seen as a sign of an anomaly.
On the other hand, consider a tag that represents nearly random event, or has values that are difficult to forecast (e.g., the atmospheric pressure in the environment).
If it reports an unusual value, it might represent a normal state of the industrial process, which corresponds to a simple fluctuation for the tag value.
This means that we should give more weight to tags that are easier to forecast and less weight to tags that are difficult to forecast.

The third technique is to use weights when calculating $M_t$:
\begin{equation}M_t^{(3)} = \frac{1}{m}\sum_{i=1}^{m}w_i|\tilde{x}_{ti} - x_{ti}|^p\end{equation}
It takes several steps to obtain the precise weight values (Algorithm ~\ref{mse_weights_calc}).

\begin{algorithm}[]
\AlgoDontDisplayBlockMarkers\SetAlgoNoEnd\SetAlgoNoLine
\caption{Weight calculcation}
\label{mse_weights_calc}
$\varepsilon_i$ = 99th percentile of prediction error for tag $i$.

$\hat{\mathcal{E}} = \max\limits_{t,i}(|\tilde{x}_{ti} - x_{ti}|)$

$\mathcal{E} = \max(\hat{\mathcal{E}}, 10^{-8})$

$\hat{\varepsilon}_i = \max(\varepsilon_i / \mathcal{E}, 10^{-8}), i = \overline{1, m}$

$\hat{w}_i = -\ln(\hat{\varepsilon}_i), i = \overline{1, m}$

$\sigma_w = \sum\limits_{i=1}^{m}\hat{w}_i, i = \overline{1, m}$

$w_i = \hat{w}_i / \sigma_w, i = \overline{1, m}$
\end{algorithm}

Note that $\hat{\varepsilon}_i \in (0; 1]$, hence $w_i > 0$ and the equality $\sum_{i=1}^{m}w_i = 1$ is always satisfied.

\subsubsection{Disjoint Forecasting Window}
The fourth technique is to forecast for a time window that is at some distance from the time window upon which the forecast is based, i.e., forecasting horizon $h > 0$ (Figure ~\ref{fig:shifted_forecast}).
For example, a forecasting model might get values from the time period from 16:00 to 16:02 and, based on those values, forecast what will happen from 16:05 to 16:06.

At first glance, this seems unnecessary, as we can try to forecast from 16:02 to 16:06 and obtain even better results.
However, there are some reasons for predicting for a time window $\tilde{W}$ that is disjoint from the input window $W$.
One reason is that the greater the forecast length $L$, the lower the prediction accuracy.
Another reason is that a models with a joint forecasting window (i.e. $h = 0$) is often prone to just copying the last values of the base window $W_k$ to $\tilde{W}_k$, working as a simple linear predictor.
And because most of the tags represent continuous value (and it won't change fast), forecast error $M_t$ won't be very high in any moment $t$.
Hence, such a system might be a bad choice for anomaly detection.
The third reason is that some anomalies take a long time and some time windows will even fully belong to that anomaly.
As a result, the forecast will be based on anomalous data and hence will not be legitimate.

\subsection{Diagnosing anomalies}
In a real-world industrial control system it is always important not only to detect an anomaly, but also identify the specific part of the system that is misbehaving.
Our detection method enables us to analyze $E_{ti} = |\tilde{x}_{ti} - x_{ti}|$ for each timepoint $t$ and tag $i$.
Note that if there is an anomaly, its location should match that of the greatest forecast error.
Thus, if $E_{ta} \geq E_{ti}$ then tag $a$ is reported as an attack target for a timepoint $t$ (assuming $M_t^{(3)} \geq T$).

Let's assume we have an ideal predictor.
The greatest error value in the prediction will point to the most erroneous (i.e. anomalous) behavior.
This method of diagnosing anomalies works quite well and the results can be seen in Table ~\ref{tab:anomaly_reason}.
Note that only two tags with the greatest prediction error are listed as a detection.

Note also that the first digit in a tag's label represents one of the six subprocesses of the testbed.
For example, in attack 12, tag LIT301 was under attack and our system detected an anomaly in tags MV303 and MV301, which are in the same part of the physical system as the attack target.

\begin{table}[h]
\caption{Example of anomaly diagnosis. A more detailed description can be found in Appendix A.}
\begin{tabular}{| c | l | l |}
    \hline
    Attack & Target & Detection \\
  \hline
1 & MV101 & - \\
2 & P102 & MV301, \textbf{P102} \\
3 & LIT101 & - \\
4 & MV504 & - \\
5 & AIT202 & \textbf{AIT202}, P203 \\
6 & LIT301 & \textbf{LIT301}, PIT502 \\
7 & DPIT301 & \textbf{DPIT301}, MV302 \\
8 & FIT401 & \textbf{FIT401}, PIT502 \\
9 & FIT401 & MV304, MV302 \\
10 & MV304 & -  \\
11 & MV303 & -  \\
12 & LIT301 & MV301, MV303 \\
13 & MV303 & - \\
14 & AIT504 & \textbf{AIT504}, P501 \\
15 & AIT504 & - \\
16 & MV101, LIT101 & UV401, P501 \\
17 & UV401, AIT502, P501 & DPIT301, MV302 \\
18 & P602, DIT301, MV302 & P302, P203 \\
19 & P203, P205 & MV101, LIT401 \\
20 & LIT401, P401 & P602, MV303 \\
21 & P101, LIT301 & LIT401, AIT402 \\
22 & P302, LIT401 & - \\
23 & P302 & MV201, LIT101 \\
24 & P201, P203, P205 & LIT401, AIT503 \\
25 & LIT101, P101, MV201 & LIT301, FIT301 \\
26 & LIT401 & P602, MV303 \\
27 & P101 & MV201, P203 \\
28 & P101, P102 & MV201, MV303 \\
29 & LIT101 & \textbf{LIT101}, AIT503 \\
30 & P501, FIT502 & FIT504, FIT503 \\
31 & AIT402, AIT502 & \textbf{AIT502}, \textbf{AIT402} \\
32 & FIT401, AIT502 & \textbf{FIT401}, P201 \\
33 & FIT401 & UV401, \textbf{FIT401} \\
34 & LIT301 & - \\
  \hline
\end{tabular}
\label{tab:anomaly_reason}
\end{table}

Some models achieve better detection score, but lower attack target accuracy.
For the SWaT system with its 51 tags, our model can detect an attack target with a 95\% accuracy if we look only at the top five highest error values ($E_{ta}$).

Note that common methods of anomaly detection like Dimensionality Reduction ~\cite{dim_reduction}, Isolation Forest ~\cite{isol_forest}, and Oneclass SVM ~\cite{one_class_svm} don't have an easy way of diagnosing anomalies.

%% file: arch_search.tex
\section{Searching for the optimal architecture}
\label{arch_search}

In most cases, a neural network designed to solve a problem is built in two stages.
First, an expert analyzes well-known architectures to decide which one of them is the most suitable for the task.
Next, after building a basic variant of the chosen architecture, the expert fine-tunes it "manually".
The second stage involves a huge amount of manual work, because it is often not obvious which configuration is the most appropriate.
As part of this study, we proposed an approach to automatically searching for suitable network architecture, which proved very helpful in the second stage of this research.

Automatic neural network generation via evolutionary computation ~\cite{gen_nn_via_ec} and genetic algorithms ~\cite{gen_nn_via_ga} has been researched before.
Most of the time neuroevolution is used to optimize both network architecture and neuron weights (e.g. ~\cite{cma_tweann}), but some researchers use it only for architecture optimization (~\cite{hier_search}) and then train networks with supervised learning methods.
We use the second approach: generate architecture via genetic algorithm and then train given network via backpropagation ~\cite{backprop}.
It is still not used very often because it has some drawbacks, such as high convergence time and excessive computational power requirements.

\subsection{Architecture Search Space}
The main idea behind our approach is that the researcher describes an architecture template (using a special description language), after which one or more neural networks are generated.
They will always satisfy the original architecture template and the best of these NN between them is selected as the solution.
Each description consists of several parts:
\begin{itemize}
\item Optimizer (and its parameters)
\item Weight and bias initializers
\item Maximum number of layers
\item For every layer we define:
\begin{itemize}
    \item A list of possible layer types, which is a subset of: Dense, Convolutional ~\cite{conv}, GRU ~\cite{gru}, LSTM ~\cite{lstm}, Dropout ~\cite{dropout}
    \item Activation function: Linear, ReLU ~\cite{relu}, Tanh, Sigmoid, Softmax.
    \item Possible layer size (as a range or a probability distribution function)
\end{itemize}
\end{itemize}

For our SWaT dataset analysis, we used three architecture templates, each representing one of the well-known approaches for a time-series analysis: multilayer perceptron, convolutional networks, recurrent neural network ~\cite{icml2017}.

\subsection{Optimization with Genetic Algorithms}
We used genetic algorithms ~\cite{ga_tutorial}  to find the best solution that satisfied the given template.
Mean squared error was used as the loss function to train each NN.
$MSE$ on a training dataset was used as an NN $fitness$ measure.
%

There are a several hyperparameters when using genetic algorithms.
We set the initial population size $N$ to 10, as enabled us to complete a enough evolution steps within a reasonable timeframe.
We found experimentally that $death age$ should not be high (we used an age of 3) or the population will stagnate, reaching the local maximum too quickly.
Using multi-parent crossover has been proved to be very efficient ~\cite{ga_convergence}.
We used $p_n = 3$, as it proved to give enough convergence speed.
The reason for not using more parents for crossover is that we need a greater population size for multi-parent crossover (otherwise, the parent pool will lack diversity).
The population size should be proportional to the square of the number of parents, which means that more calculations are needed to complete the same number of evolution steps.

\subsection{Mutation and Crossover Operations}
A mutation operation is performed in several steps (Algorithm ~\ref{mutation_process}).

\begin{algorithm}[]
\AlgoDontDisplayBlockMarkers\SetAlgoNoEnd\SetAlgoNoLine
\caption{Mutation Process}
\label{mutation_process}

\Fn{Mutate($NN$, $D$)}{
\KwIn{$NN$ - original neural network,

$D$ - architecture description}
\KwOut{$MNN$ - mutated neural network}
$MNN$   $\gets$ copy($NN$)\;
$i$ $\gets$ RandomLayerIndex($MNN$)\;
Using $D$, generate new layer $i$ configuration\;
\Return $M$\;
}
\end{algorithm}
\noindent

A crossover operation is performed layer-by-layer.
It is a bit tricky as the architecture description has two types of parameters: categorical (optimizer and layer types, activation functions, etc.) and numeric (learning rates, layer sizes, etc.).
For each of the categorical parameters we look at the parent values and use the most frequently occurring.
In the event of a tie, we choose randomly.
For example, if activation functions for the second layer of parents are 'relu', 'linear', 'relu', then the descendant will have 'relu' as its activation function for the second layer.
If activation functions for the third layer of parents are 'relu', 'sigmoid', 'linear', then the choice will be made randomly from the three values.

Crossover of numeric values is performed as binary occurrence-based scanning ~\cite{ga_convergence} operation.
Technically, the bitwise vote function is used to calculate the resulting value.
Let $a$, $b$, $c$ be the parent values.
Then the descendant value $d = (a \land b) \lor (a \land c) \lor (b \land c)$.
All operations are bitwise.
For example, crossover between numbers 12 (binary 1100), 5 (binary 0101) and 15 (binary 1111) will be 13 (binary 1101).

\subsection{Best Performing Model}
The architecture of the best model found is described below.
It uses an input window of length $L = 200$, forecast horizon $h = 50$ and forecast length $\tilde{L} = 4$.

We used Keras ~\cite{chollet2015keras} library to train our NN. 
The best-scoring neural network architecture is shown on Table ~\ref{tab:arch}.

\begin{table}
\caption{Architecture of the best performing model found during the process of NN architecture optimization.}
\begin{tabular}{ | c | c | c | c | }
    \hline
    Layer & Type & Role & Output Shape \\
  \hline
1 & Dense & Encoder & (51, 43) \\
2 & Dense & Analyzer & 96 \\
3 & Dense & Analyzer & 71 \\
4 & Dense & Decoder & (4, 51) \\
    \hline
\end{tabular}
\label{tab:arch}
\end{table}

It was trained using the Adam ~\cite{adam} optimizer with default parameters and default initializers.
The entire dataset consisted of 105 527 samples.

%% file: detection_metrics.tex
\section{Detection Quality Metrics}
\label{detection_metrics}

To tell which of the anomaly detection methods performs better, one needs a metric.
Some studies ~\cite{swat_detect} used the $F_1$ score as a metric.
However, in this study, we encountered a very unintuitive behavior of the $F_1$ score and used $NAB$ score too ~\cite{NAB}.
An ideal detector will achieve a $NAB$ score of 100 and a detector with zero detections (\"null-detector\") will achieve 0 $NAB$-score.
It is rare, but possible to achieve negative $NAB$ score in the case when sum of false positive and false negative rates is greater then true-positive rate.
Our best model based on a multilayer perceptron (MLP) achieved an $F_1$ score of 0.812.
It also produced a 69.612 $NAB$ score.

We expected that ignoring all the periodic tags (31 out of 51) would significantly harm the detection system.
But when we trained our best model on a "shrinked" dataset (i.e. data without any nearly periodic series), it produced an $F_1$ score of 0.767, which is only 6\% worse than our best result.
There were 14 attacks (out of 34 with physical impact) in the dataset that targeted periodic processes.
Although ignoring the 31 periodic tags significantly affected the detection quality (since the model nearly ignored 14 attacks), the $F_1$ metric failed to reflect it.

The reason for such strange behavior lies in the nature of the $F_1$ calculation.
It focuses on the duration of all anomalies detected.
Thus, if we detect one anomaly that is one hour long and skip 10 anomalies of one minute each, then we achieve a precision of 0.86, although we detected only one anomaly out of eleven.
And that's the case of the SWaT dataset: attack 28 is about 10 hours long, which is 65\% of the total time for all the anomalies in the SWaT dataset.
Hence for the $F_1$ score about 65\% of the model's success will depend on detecting a single (and easily perceptible) anomaly.

On the other hand, the $NAB$ metric scores anomalies regardless of their duration, it emphasizes the precise moment of detection.
Thus, a model that detects all of the anomalies with a huge delay will score far fewer points than a model that detects the same anomalies earlier.
Notably, for a real-world ICS early detection is crucial.

\begin{table}
\caption{Comparison with state-of-the-art models. Results for SVM and DNN models were taken from ~\cite{SWaT_dataset}.}
\begin{tabular}{| l | c | c | c | c |}
\hline
Model & NAB score & $F_1$ & Precision & Recall\\
\hline
MLP & \textbf{69.612} & \textbf{0.812} & 0.967 & 0.696 \\
CNN & 34.225 & 0.808 & 0.952 & \textbf{0.702} \\
RNN & 36.924 & 0.796 & 0.936 & 0.692 \\
\hline
SVM* & - & 0.796 & 0.925 & 0.699 \\
DNN* & - & 0.802 & 0.982 & 0.678 \\
\hline
\end{tabular}
\label{tab:scores}
\end{table}

If the Table ~\ref{tab:scores} we compare the results of our three best models with the state-of-the-art solutions for the SWaT dataset.
Our models are based on three well-known neural network templates: multilayer perceptron (MLP), convolutional network (CNN) and recurrent network (GRU-based).
Notably, network based on GRU units performed better then network, based on LSTM units.

Another example is a model trained with a shifted forecast window.
We trained a model that made its forecast on $L = 200$ data points (as mentioned above, each data point represents one second) and predicted $\tilde{L} = 4$ values with forecast horizon $h = 50$.
For example, based on tag values from 16:00:00 till 16:03:20, it predicted values for the time interval from 16:04:10 to 16:04:13.
The model had good detection capabilities (18 true positives and a $NAB$ score of 45.196), but got an $F_1$ score of 0.746.
Note that the model based on a "shrinked" dataset scored better in the $F_1$ metric.

The proposed solution to such unwanted behavior is to use the $NAB$ score, since it rewards early detection and is not affected by the anomaly duration.

%% file: conclusions.tex
\section{Conclusions}
\label{conclusions}

In this paper, we further  developed NN-based forecasting approach to early anomaly detection. 

A method for network architecture optimization with genetic algorithm was presented.

We used the SWaT dataset, $NAB$-score and $F_1$ score to evaluate the quality of the developed detection model.
The best NN detects 25 anomalies out of 34 and shows 7 false positives (see Appendix A).
It got a $NAB$-score of 69.612 and an $F_1$ score of 0.812, with a precision of 0.976 and recall of 0.696.
The average delay for anomaly detection is 11\% of the anomaly's length.

Most calculations were done using the following hardware: Intel(R) Xeon(R) CPU E5-2620 v3 @ 2.40GHz, 64 GB RAM, nVidia Tesla P4. It took 33 hours of machine time to find the best architecture and five minutes (with a batch size 2000) to train corresponding NN.

It was also shown that anomaly detection researchers should use the $F_1$ score with care as it's prone to overscore detection of long anomalies and underscore detection of short anomalies.
In a real ICS even short-time anomaly (e.g. nuclear reactor fault) might have devastating consequences.

We proposed several techniques to improve AD quality: exponentially weighted smoothing to decrease false-positive rate, mean p-powered error measure to decrease false-negative rate, individual error weight for each variable to reduce sensitivity for noisy variables, disjoint prediction windows to avoid using last values a prediction.  

One of the main features of the approach presented is its ability to interpret an anomaly, or, in terms of an ICS, locate the sensor (or sensors) being attacked.
For each attack the detection system will report the most suspicious sensors with an accuracy of 95\%.

%% file: acknowledgement.tex
\section{Acknowledgements}
\label{acknowledgement}

The authors are sincerely grateful to Artem Vorontsov for useful discussions.
The authors want to thank iTrust, Centre for Research in Cyber Security, Singapore University of Technology and Design for presenting the SWaT dataset.
This work was supported by the Kaspersky Lab.

%% file: appendix_a.tex
\onecolumn
\section{Detailed Model Comparison}

\begin{table}[H]
\caption{Comparison of different detection models. Results for SVM and DNN models are taken from ~\cite{SWaT_dataset}.}
\begin{tabularx}
\textwidth{| c | l | l | l | X | X | X |}
\hline
Attack & Target & Detection & Detection Delay (s) &  \multicolumn{3}{c |}{Recall} \\
\hline
& & & & MLP & DNN* & SVM* \\
\hline
1 & MV101 & - & - & 0 & 0 & 0 \\
2 & P102 & MV301, \textbf{P102} & 112 & \textbf{0.764} & 0 & 0\\
3 & LIT101 & - & - & 0 & 0 & 0 \\
4 & MV504 & - & - & 0 & 0 & \textbf{0.035} \\
5 & AIT202 & \textbf{AIT202}, P203, & 11 & \textbf{0.952} & 0.717 & 0.720\\
6 & LIT301 & \textbf{LIT301}, PIT502, & 42 & \textbf{0.909} & 0 & 0.888\\
7 & DPIT301 & \textbf{DPIT301}, MV302, & 16 & \textbf{0.984} & 0.927 & 0.919\\
8 & FIT401 & \textbf{FIT401}, PIT502, & 18 & 0.976 & \textbf{1} & 0.433\\
9 & FIT401 & MV304, MV302, & 3 & 0.989 & 0.978 & \textbf{1}\\
10 & MV304 & - & - & 0 & 0 & 0\\
11 & MV303 & - & - & 0 & 0 & 0\\
12 & LIT301 & MV301, MV303, & 297 & \textbf{0.603} & 0 & 0\\
13 & MV303 & - & - & 0 & 0 & 0 \\
14 & AIT504 & \textbf{AIT504}, P501, & 13 & \textbf{0.97} & 0.123 & 0.130 \\
15 & AIT504 & - & - & 0 & 0.845 & \textbf{0.848} \\
16 & MV101, LIT101 & UV401, P501, & 10 & \textbf{0.98} & 0 & 0.016\\
17 & UV401, AIT502, P501 & DPIT301, MV302, & 16 & 0.978 & 0.998 & \textbf{1}\\
18 & P602, DIT301, MV302 & P302, P203, & 102 & 0.711 & \textbf{0.876} & 0.875\\
19 & P203, P205 & MV101, LIT401, & 53 & \textbf{0.918} & 0 & 0\\
20 & LIT401, P401 & P602, MV303, & 1043 & \textbf{0.294} & 0 & 0.009\\
21 & P101, LIT301 & LIT401, AIT402, & 67 & \textbf{0.998} & 0 & 0\\
22 & P302, LIT401 & - & - & 0 & 0 & 0 \\
23 & P302 & MV201, LIT101, & 1164 & 0.0324 & \textbf{0.936} & 0.936 \\
24 & P201, P203, P205 & LIT401, AIT503, & 52 & \textbf{0.87} & 0 & 0\\
25 & LIT101, P101, MV201 & LIT301, FIT301, & 105 & \textbf{0.834} & 0 & 0.003\\
26 & LIT401 & P602, MV303, & 102 & \textbf{0.786} & 0 & 0 \\
27 & P101 & MV201, P203, & 89 & \textbf{0.331} & 0 & 0 \\
28 & P101, P102 & MV201, MV303, & 82 & \textbf{0.84} & 0 & 0 \\
29 & LIT101 & \textbf{LIT101}, AIT503, & 110 & \textbf{0.808} & 0 & 0.119\\
30 & P501, FIT502 & FIT504, FIT503, & 79 & 0.842 & \textbf{1} & 1\\
31 & AIT402, AIT502 & \textbf{AIT502}, \textbf{AIT402}, & 73 & 0.767 & 0.923 & \textbf{0.927}\\
32 & FIT401, AIT502 & \textbf{FIT401}, P201, & 71 & 0.836 & \textbf{0.940} & 0\\
33 & FIT401 & UV401, \textbf{FIT401}, & 71 & 0.784 & \textbf{0.933} & 0.927 \\
34 & LIT301 & - & - & 0 & 0 & \textbf{0.357} \\
\hline
\end{tabularx}
\label{tab:detects}
\end{table}